\documentclass[journal,twoside,web]{IEEEtran}
\usepackage{amsmath,amssymb,amsfonts}
\usepackage{algorithmic}
\usepackage{graphicx}
\usepackage{algorithm,algorithmic}
\usepackage{adjustbox}
\usepackage{threeparttable}
\usepackage{booktabs}
\usepackage{multirow}
\usepackage{tabularx}
\usepackage{subfigure}
\usepackage{hyperref}
    
\hypersetup{hidelinks=true}
\usepackage{textcomp}

\def\BibTeX{{\rm B\kern-.05em{\sc i\kern-.025em b}\kern-.08em
    T\kern-.1667em\lower.7ex\hbox{E}\kern-.125emX}}
\begin{document}
\title{EFCM: Efficient Fine-tuning on Compressed Models for deployment of large models in medical image analysis}
\author{Shaojie Li, Zhaoshuo Diao
\thanks{This work was supported by the National Natural Science Foundation of China (No.********). (Corresponding author: ******.)}
\thanks{Shaojie Li is with Zhejiang Lab, Hangzhou 311121, China (e-mail: lyconan126@163.com)}
\thanks{Zhaoshuo  Diao is with the School of Software, Shenyang University of Technology, Shenyang 110870, China (e-mail: zsdiao@sut.edu.cn)}
}

\maketitle

\begin{abstract}
The recent development of deep learning large models in medicine shows remarkable performance in medical image analysis and diagnosis, but their large number of parameters causes memory and inference latency challenges. Knowledge distillation offers a solution, but the slide-level gradients cannot be backpropagated for student model updates due to high-resolution pathological images and slide-level labels. This study presents an Efficient Fine-tuning on Compressed Models (EFCM) framework with two stages: unsupervised feature distillation and fine-tuning. In the distillation stage, Feature Projection Distillation (FPD) is proposed with a TransScan module for adaptive receptive field adjustment to enhance the knowledge absorption capability of the student model. In the slide-level fine-tuning stage, three strategies (Reuse CLAM, Retrain CLAM, and End2end Train CLAM (ETC)) are compared. Experiments are conducted on 11 downstream datasets related to three large medical models: RETFound for retina, MRM for chest X-ray, and BROW for histopathology. The experimental results demonstrate that the EFCM framework significantly improves accuracy and efficiency in handling slide-level pathological image problems, effectively addressing the challenges of deploying large medical models. Specifically, it achieves a 4.33\% increase in ACC and a 5.2\% increase in AUC compared to the large model BROW on the TCGA-NSCLC and TCGA-BRCA datasets. The analysis of model inference efficiency highlights the high efficiency of the distillation fine-tuning method.
\end{abstract}


\section{Introduction}
\label{sec:introduction}
\IEEEPARstart{R}{ecently}, deep learning models have emerged as potent tools in medicine. They have shown outstanding performance in medical image analysis \cite{li2023medical}, disease diagnosis, and treatment planning. The emergence of large models has further promoted the application of deep learning in the medical field. In medical image processing, large models achieve more accurate feature extraction and analysis, more accurate disease diagnosis and classification, as well as better understanding and processing capabilities for complex pathological images, thus providing a more reliable basis for medical diagnosis and treatment, so large models have great application value and development potential in the medical field. The large model Virchow proposed by Vorontsov \textit{et al.} \cite{vorontsov2023virchow}, has 632 million parameters and surpasses state-of-the-art methods across multiple computational pathology tasks.

However, despite the remarkable achievements and potential of large models in medicine, the huge number of model parameters makes it challenging to deploy these models online or on mobile devices in terms of memory cost and inference latency \cite{zhang2023challenges}. 

In recent years, knowledge distillation has emerged as a promising approach for training lightweight deep neural network models in computer vision tasks \cite{hao2022cdfkd}. The core idea behind knowledge distillation is to train a compact student model to mimic the outputs, or soft labels, of a pretrained cumbersome teacher model. This method is initially introduced by Hinton \textit{et al.} \cite{hinton2015distilling}. However, existing distillation methods have limitations when dealing with slide-level pathology images. Pathology images usually have huge resolution and are only available with slide-level label \cite{lu2021data}. To deal with this situation, it is usually necessary to segment the whole slide image (WSI) into small instances and use a Multiple Instance Learning (MIL) \cite{ilse2018attention} approach to synthesize a series of instances as a bag of samples for decision-making. However, end-to-end training on the MIL classification problem is very difficult due to the computational limitations, as the slide-level gradients cannot be backpropagated in parallel to a feature encoder with more than 10k instances of a bag \cite{pisula2024fine}.

Also, typically many large models use transformer architectures, and we need the student model to be small enough. If the student model is also a transformer architecture, although they can align features in the same feature space, it is very challenging to transfer extensive knowledge from a large model with hundreds of millions of parameters to a tiny model with millions of parameters by distillation \cite{kanwal2023vision}. Thus, how to improve the knowledge absorption of the student model has become an urgent problem.

In the domain of unsupervised domain adaptation, Liang \textit{et al.} \cite{liang2021distill} proposed a distill and fine-tune two-step adaptive framework, which has been demonstrated to be effective. To address the first problem, we propose a distillation followed by fine-tuning approach as the framework of Efficient Fine-tuning on Compressed Models (EFCM). First, a compact student model is trained on feature dimensions using the unsupervised feature distillation technique in knowledge distillation. Then, we further optimize the distilled student model using an end-to-end fine-tuning strategy.

In order to enhance the knowledge absorption of the student model, inspired by the proposal of \cite{xiao2021early} and \cite{li2019selective}, we propose the Feature Projection Distillation (FPD) method. For the neurons to capture targets at different scales, we propose a novel TransScan module, which mainly consists of the transformer and the Selective Convolutional Attention Network (SCAN). The SCAN achieves adaptive tuning of receptive field size through a selective convolution mechanism, thus improving the model's knowledge absorption ability.

In summary, the EFCM framework provides a novel solution to the challenges of deploying large-scale models in the medical domain. It brings significant advantages to the field of medical image analysis in terms of optimizing computational cost, memory cost, and inference latency. And it opens up new opportunities for the application of large-scale models in the medical field.

The main contributions of this work are as follows:
\begin{itemize}
\item[$\bullet$]We construct the EFCM framework. By applying the unsupervised feature distillation technique to distill the large model, and adopting End2end Train CLAM (ETC), a fine-tuning strategy for the distilled student model, the model efficiency and performance are significantly improved in dealing with the slide-level pathology image classification problem.
\item[$\bullet$]We also propose an FPD method, which uses the selective convolution mechanism introduced in the TransScan module to achieve adaptive adjustment of the receptive field size, and adopts Mean Squared Error (MSE) and Kullback-Leibler (KL) divergence as the distillation loss to further enhance the model.
\item[$\bullet$]We analyze the full-parameter fine-tuning, parameter-efficient fine-tuning, and distillation fine-tuning methods in terms of inference metrics such as parameters (Params), Memory Access Cost (MAC), Giga Floating-point Operations Per Second (GFLOPS), and Frames Per Second (FPS), highlighting the high efficiency of the distillation fine-tuning methods.
\end{itemize}

\section{Related Work}
\label{sec:rel}
In this section, we present a concise review of the existing literature, focusing on three key areas: medical large models, knowledge distillation and fine-tuning.

\subsection{Medical Large Models}
In recent years, the field of large medical models has been booming. Large models have demonstrated great adaptability and versatility. Zhang \textit{et al.} \cite{zhang2023biomedgpt} introduce BiomedGPT, which can perform a variety of tasks in the biomedical domain across multiple modalities (\textit{e.g.}, radiographs, digital images, and text). Wu \textit{et al.} \cite{wu2023towards} introduce the Radiological Fundamental Model (RadFM), which effectively fuses medical scans with natural language, demonstrating the advantages of RadFM in visual and textual information synthesis. Chen \textit{et al.} \cite{chen2023general} propose UNI, a large-scale pathology model based on self-supervised learning that outperforms previous techniques in various computational pathology tasks. However, deploying large models remains challenging due to the black-box nature of many models (accessible via APIs) and their high computational cost. Hence, alternative solutions are needed to harness the capabilities of large models for knowledge-intensive inference tasks.

\subsection{Knowledge Distillation}
Knowledge distillation is an effective approach for compressing models, leveraging the output logits of a pre-trained teacher model as guidance to train lightweight student models. This concept is initially proposed by Buciluǎ \textit{et al.} \cite{buciluǎ2006model} and further refined by Hinton \textit{et al.} \cite{hinton2015distilling}. Subsequent work further improves logits-based knowledge distillation through structural information, model ensembling, or contrastive learning. Recently, Huang \textit{et al.} \cite{huang2022knowledge} introduce a distillation approach that relaxes the KL divergence loss to accommodate significant capacity disparities between teacher and student. Apart from logits, some knowledge distillation methods utilize intermediate features as hints. Yim \textit{et al.} \cite{yim2017gift} employ flow-based process matrices generated from features as hint knowledge. Additionally, there are numerous other feature distillation methods utilizing various hint designs \cite{chen2021distilling}.
\begin{figure*}[!t]
    \centering
    \includegraphics[scale=.15]{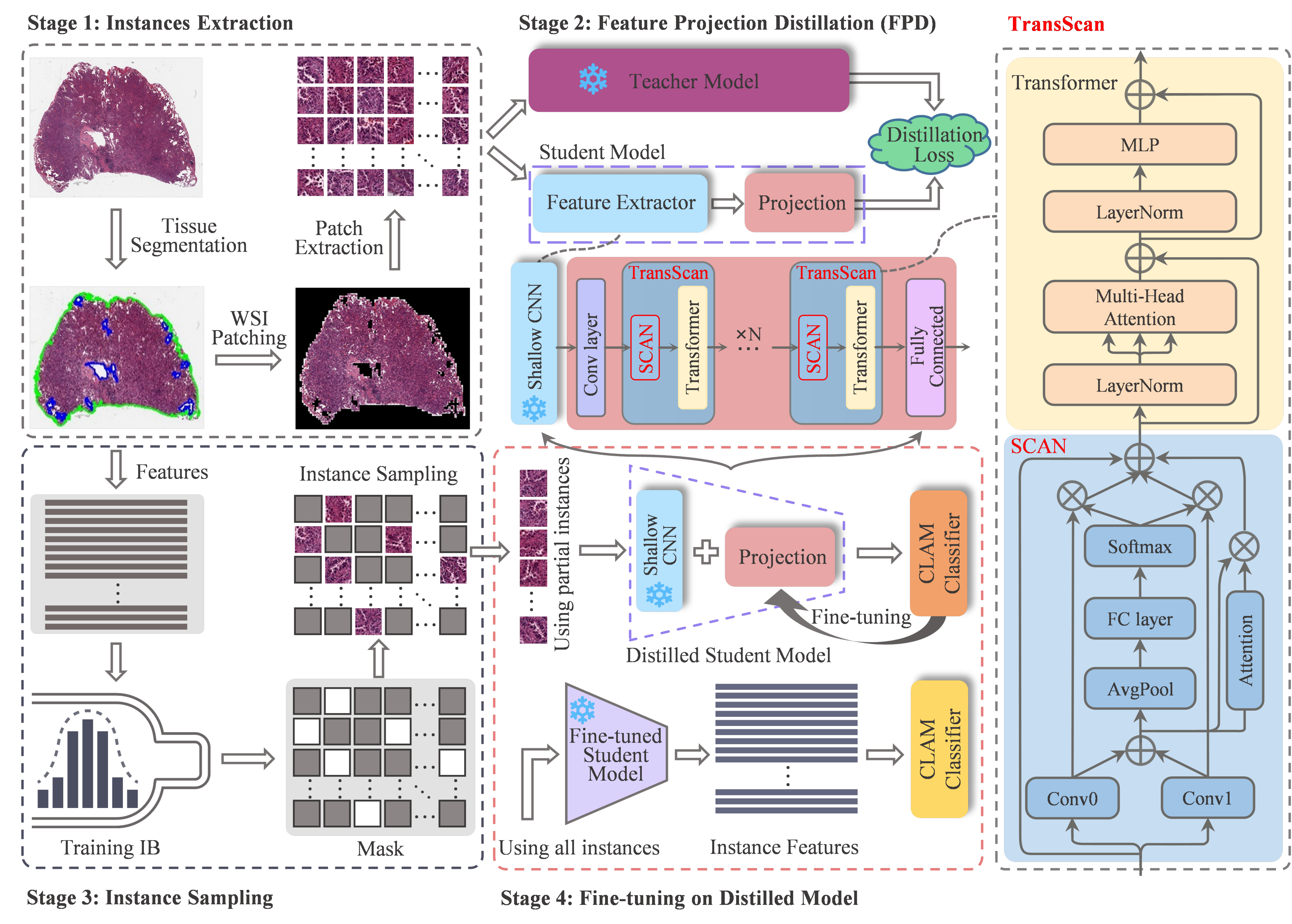}
    \caption{The framework of EFCM for slide-level pathology images. Stage 1: Extract tissue regions from the WSI and perform patch extraction within these regions. Stage 2: Utilize a large pre-trained model as the teacher model to guide knowledge transfer to the student model through distillation. Stage 3: Employ instance features extracted by the teacher model to train the Information Bottleneck (IB) module for generating instance masks, filtering a restricted number of instance samples per WSI. Stage 4: Fine-tune the distilled student model end-to-end, and then use the fine-tuned student model as a feature extractor to extract features from all instance samples to further train a new CLAM classifier.
    }
    \label{fig:framework}
\end{figure*}

Despite the significant performance improvement achieved by existing feature-based distillation methods, most of them use feature hints as an auxiliary to guide output prediction. However, when faced with slide-level pathological image classification, the slide-level gradients cannot be backpropagated to a feature encoder in parallel due to computational limitations.

\subsection{Fine-Tuning}
When fine-tuning the entire network for downstream tasks, the exponential growth of model parameters poses computational challenges, and full-parameter fine-tuning may result in a decrease in the out-of-distribution (OOD) performance of pre-trained models \cite{kumar2022fine}. Consequently, some researchers explore parameter-efficient fine-tuning methods to train subsets of the model or add modules with fewer parameters while achieving comparable or even superior performance. Methods like Adapter \cite{chen2022adaptformer} insert trainable modules (\textit{e.g.}, Multilayer Perceptrons (MLPs) with activation functions and residual structures) into the network to facilitate transfer learning. LoRA \cite{hu2021lora} leverages low-rank updates to large-scale frozen models and introduces bypass paths to mimic fine-tuning of the entire model parameters. Despite some success achieved by LoRA and Adapter methods, there may be limitations in applicability and performance on specific tasks or datasets. There is a trade-off between compression and performance preservation in these methods. To further improve model compression, we propose to use feature distillation techniques to compress pre-trained models into smaller models while maintaining performance during the fine-tuning process. 

\section{Method}
This section presents a novel EFCM framework that addresses the limitation that pathology image MIL classification cannot effectively backpropagate the sliding gradient to update the feature encoder parameters during end-to-end training through a two-step process of distillation and fine-tuning. In the distillation stage, we propose the FPD method and adaptively adjust the receptive field size using the TransScan module. The fine-tuning is divided into slide-level and patch-level, and a progressive approach is used to compare three strategies to evaluate the performance of the distilled model: Reuse CLAM, Retrain CLAM, and End2end Train CLAM (ETC).

\subsection{Framework of EFCM}
The framework of EFCM designed for slide-level pathological images encompasses 4-stage processes. The flow of the framework is shown in Fig. \ref{fig:framework} and each stage is described below:

Initially, For slide-level histopathology images, due to their large size, we carry out preprocessing according to the operations in CLAM \cite{lu2021data}, which involves utilizing various techniques such as HSV, Blur, Threshold, and Contours to identify the tissue regions in each WSI. After identifying the tissue regions, we extract non-overlapping patches with a size of 256 \(\times\) 256, usually at a magnification of 20\(\times\) or 40\(\times\).

This is followed by a feature projection distillation stage, which utilizes a large pre-trained model to act as a teacher. The distillation mechanism is used to facilitate knowledge transfer to the student model. The student model consists of two main components, the feature extractor and the projection, the design of which is described in detail in Section \ref{sec:fpd} FPD.

Next, we extract instance features from the training set based on the teacher model for training the Information Bottleneck (IB) module. This module acts to obtain a limited number of instance samples from each WSI \cite{li2023task}. This is done to perform end-to-end fine-tuning of the distilled student model during the fine-tuning stage.

Finally, the distilled student model is fine-tuned end-to-end using partial instances. The fine-tuned student model is used as a feature extractor to extract features from all instance samples. The features are further used to train a new CLAM classifier, eventually forming a powerful and effective classification model for slide-level pathology images.

\subsection{Feature Projection Distillation (FPD)}
\label{sec:fpd}
In this section, we provide a detailed exposition of the design of the student model, the TransScan module, and the distillation loss in the FPD method. Specifically, the student model in the FPD method comprises two components: the feature extractor and the projection, with the TransScan module playing a crucial role within the projection component. The distillation loss serves as a supervisory mechanism to facilitate the alignment of predicted features generated by the student model with those of the teacher model within the feature space.

\begin{figure}[tbp]
\centering
\subfigure[Vanilla Feature Distillation]{ 
\includegraphics[scale=.16]{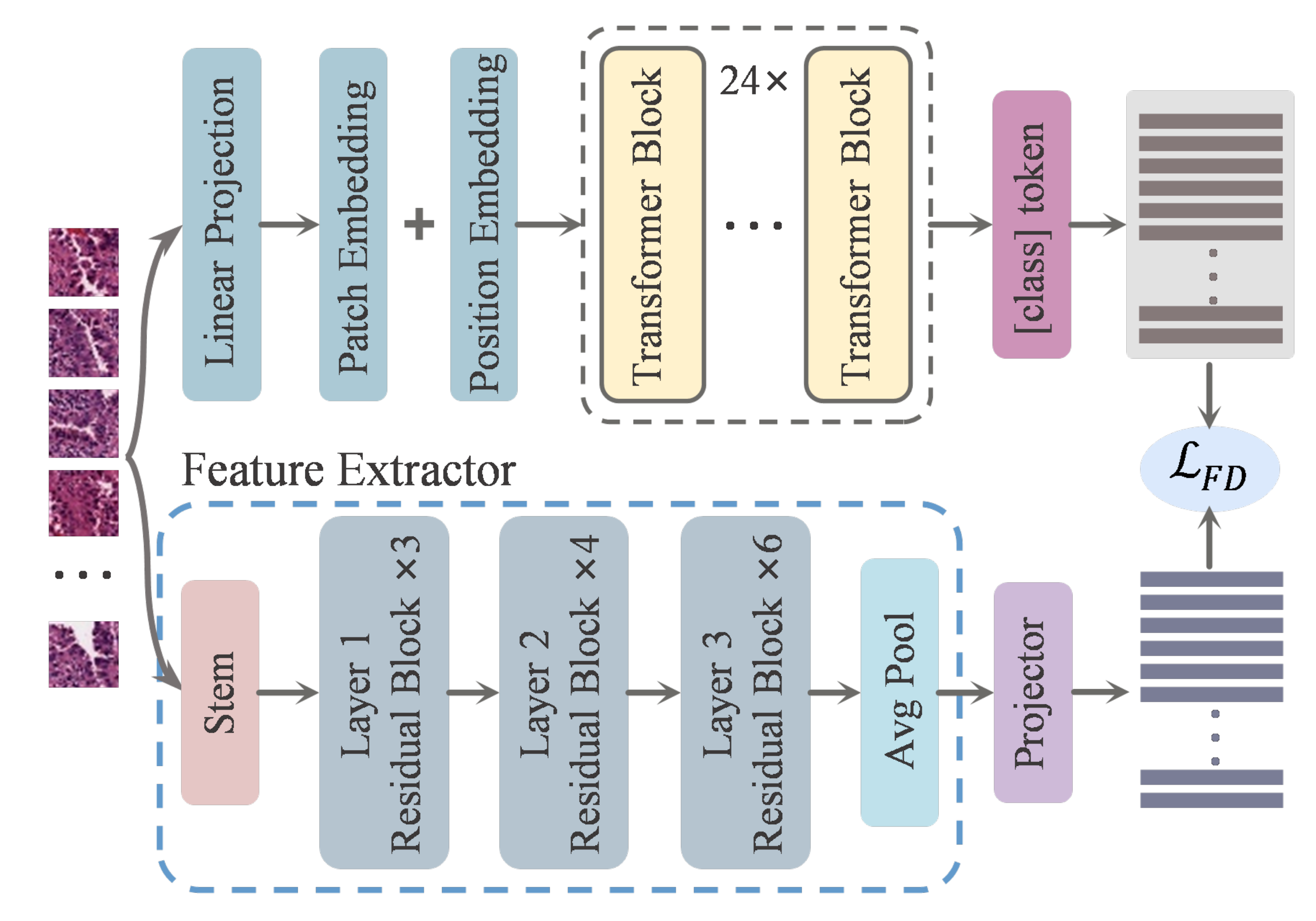}
\label{fig:fpd_a}}
\qquad
\subfigure[Feature Projection Distillation]{
\includegraphics[scale=.16]{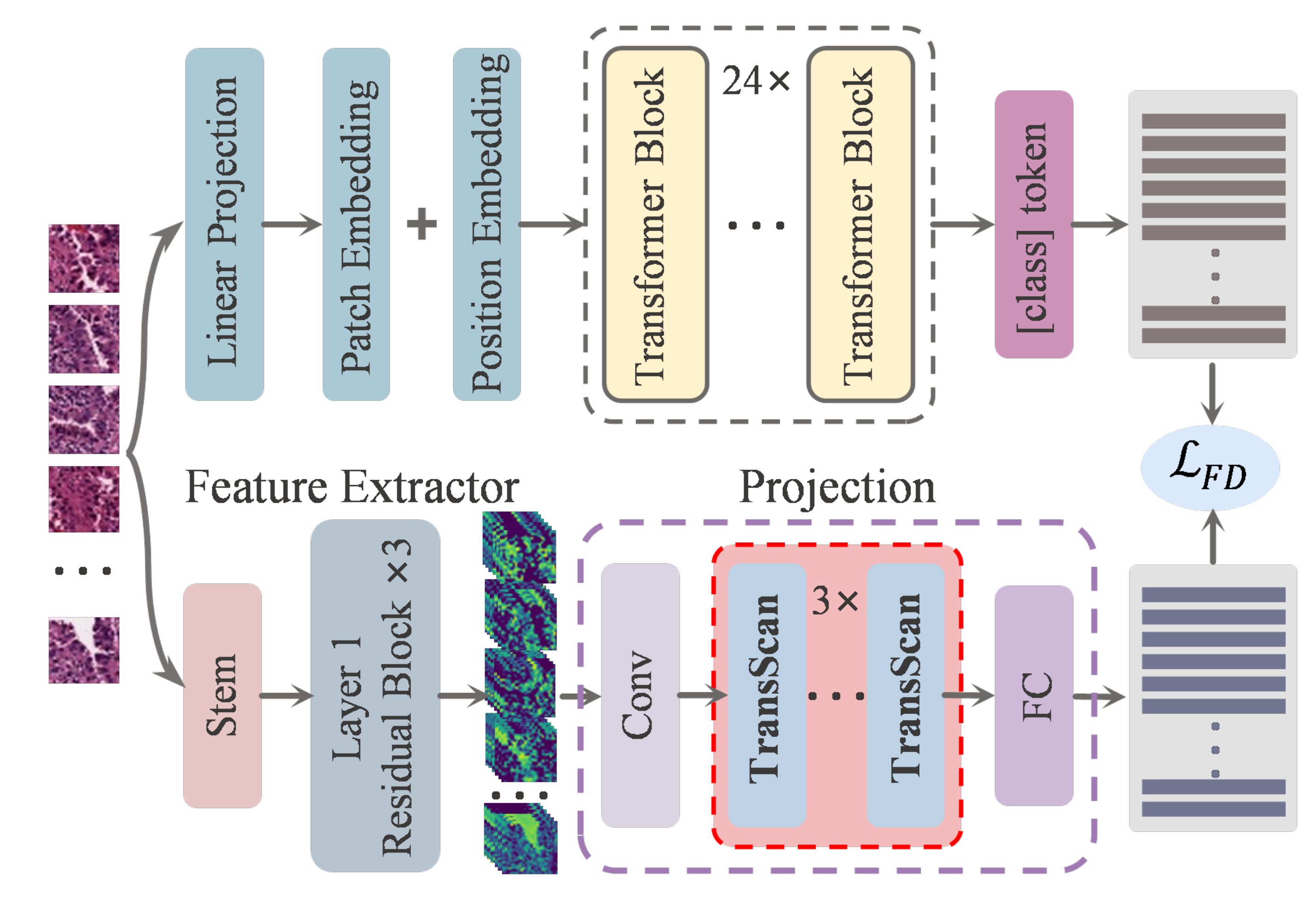}
\label{fig:fpd_b}}
\caption{Comparison of Vanilla Feature Distillation (VFD) and Feature Projection Distillation (FPD). The main differences are in the student model design and how the student model parameters are updated. (a) In VFD, the student model parameters are updated collectively. (b) In our FPD, we freeze the shallow CNN and solely update only the projection parameters. \label{fig:fpd_a_b}}
\end{figure}

\subsubsection{Design of Student Model}
The original intention of FPD design is to obtain a student model with strong knowledge absorption ability in the distillation framework. We start with conventional feature distillation and improve the design of the student model. First, we retain only the shallow CNN in Vanilla Feature Distillation (VFD) as the feature extractor. Then, we construct a projection head, mainly consisting of the TransScan module. We compare the VFD and FPD methods, as shown in Fig. \ref{fig:fpd_a_b}.

The VFD uses the soft goals generated by the teacher model to guide the training of the student model. As shown in Fig. \ref{fig:fpd_a}, the training image \(\mathbf{x}\) passes through the teacher and student models, producing the corresponding teacher feature map \({F}_{t}\) and student feature map \({F}_{s}\). Typically, differences in size and dimension between the student and teacher feature maps require the use of a projector module, usually a convolutional layer, to align them before distillation. The student model in VFD extracts features from the ``layer3" of a pre-trained ResNet50 network and aligns them directly with those generated by the teacher model through a projector function \(\mathit{\Phi}_{s}\left ( \cdot  \right )\). The projector function \(\mathit{\Phi}_{s}\left ( \cdot  \right )\) is usually a fully connected layer. The aligned student feature map \({F}_{s}\) and teacher feature map \({F}_{t}\) are represented as follows:
\begin{equation}
VFD \left\{ \begin{array}{l}
{F_{t} = \mathcal{F}_{teacher}\left( \mathbf{x} \right)} \\
{F_{s} = \mathit{\Phi}_{s} \left( {{ResNet50}_{layer3}\left( \mathbf{x} \right)} \right)}
\end{array} \right.
\end{equation}

The FPD method aims to improve the characterization of the student model. As shown in Fig. \ref{fig:fpd_b}, the FPD method mainly includes the following elements. The aligned student feature map \({F}_{s}\) and teacher feature map \({F}_{t}\) are represented as follows:
\begin{equation}
FPD \left\{ \begin{array}{l}
{F_{t} = \mathcal{F}_{teacher}\left( \mathbf{x} \right)} \\
{F_{s} = P\left( {\mathcal{F}_{n}\left( {\mathcal{F}_{2D} \left( {{ResNet50}_{layer1}\left( \mathbf{x} \right)} \right)} \right)} \right)}
\end{array} \right.
\end{equation}
Initially, we employ the shallow layer of a pre-trained ResNet50 network \cite{he2016Deep} trained on the ImageNet dataset \cite{2015ImageNet} as a feature extractor. Following this, we append a projection comprised of a 2D convolutional layer \(\mathcal{F}_{2D}\left ( \cdot  \right )\), multiple TransScans \(\mathcal{F}_{n}\left ( \cdot  \right )\) and a fully connected layer \(P\left ( \cdot  \right )\). This design aims to ensure that the student model can extract shallow feature representations from raw input data and project these features into a higher-dimensional representation space to predict features more accurately.

Specifically, the feature extraction part is ``layer1"  of a pre-trained ResNet50 network. The projection part starts with a 2D convolutional layer kernel size of 4, a stride of 4, and the input feature dimensions converted from 256 to ``dim", which is typically set to 384. The cascading TransScan module is then set to a depth of 3. Finally, the generated features are aligned to the teacher model features using a fully connected layer \(P\left ( \cdot  \right )\).

\subsubsection{TransScan Module}
\label{sec:scan}
The TransScan module comprises two key components: the transformer and the SCAN structure. The detailed structural configuration of this module is illustrated in the rightmost part of Fig. \ref{fig:framework}.

For a given input image with dimensions of \(3 \times H \times W\), after feature extraction using a shallow ResNet50, the output feature map has a size of \(256 \times \frac{H}{4} \times \frac{W}{4}\). Subsequently, the feature map undergoes 2D convolutional layer processing, resulting in a feature map \(\mathbf{X}\) with dimensions of \(``dim" \times \frac{H}{16} \times \frac{W}{16}\). For ease of subsequent presentation, we label the dimensional size of the feature map \(\mathbf{X}\) as \(C \times H^{'} \times W^{'}\). This feature map serves as the input to the TransScan module.

For the given feature map \(\mathbf{X}\) with dimensions \(C \times H^{'} \times W^{'}\), we first apply two transformations: \(\left. \tilde{\mathcal{F}}:\mathbf{X}\rightarrow\tilde{\mathbf{U}} \right.\) and \(\left. \hat{\mathcal{F}}:\mathbf{X}\rightarrow\hat{\mathbf{U}} \right.\). Specifically, both \(\tilde{\mathcal{F}}\) and \(\hat{\mathcal{F}}\) use convolution operations with a kernel size of 3, and the group number G is often set to 32. However, they differ in their padding and dilation settings, which have values of 1 and 2, respectively. Afterwards, the output feature maps from these branches are combined to generate a global feature representation:
\begin{equation}
\mathbf{U} = \tilde{\mathbf{U}} + \hat{\mathbf{U}}.
\end{equation}
To incorporate the global information, we utilize global average pooling to generate channel-wise statistics denoted as \(\mathbf{s} \in \mathbb{R}^{C}\). More specifically, the \(\mathit{c}\)-th element \(\mathbf{s}_\mathit{c}\) in \(\mathbf{s}\) is computed by spatially shrinking \(\mathbf{U}\) through spatial dimensions \(\mathit{H^{'}} \times \mathit{W^{'}}\):
\begin{equation}
\mathbf{s}_{c} = \mathcal{F}_{gap}\left( \mathbf{U}_{c} \right) = \frac{1}{H^{'} \times W^{'}}{\sum\limits_{i = 1}^{H^{'}}{\sum\limits_{j = 1}^{W^{'}}{\mathbf{U}_{c}\left( {i,j} \right)}}}.
\end{equation}
Furthermore, a concise feature vector \(\mathbf{z} \in \mathbb{R}^{d}\) is created to enable the guidance for the precise and adaptive selections:
\begin{equation}
\mathbf{z} = \delta\left( {\mathcal{B}\left( {\mathcal{F}_{fc}\left( \mathbf{s} \right)} \right)} \right),
\end{equation}
where \(\delta\) represents the ReLU function \cite{nair2010rectified}, \(\mathcal{B}\) denotes Batch Normalization \cite{ioffe2015batch}, and \(\mathcal{F}_{fc}\) symbolizes a \(1 \times 1\) convolution operation. Generally, the number of channels is made to be reduced to \(\mathit{d}\), which is frequently selected as 32. A soft attention mechanism operates across channels, directed by compact feature descriptor \(\mathbf{z}\), dynamically selecting spatial scales. Channel-wise digits undergo softmax operation using attention vectors \(\mathbf{a}\) and \(\mathbf{b}\) for \(\tilde{\mathbf{U}}\) and \(\hat{\mathbf{U}}\) respectively:
\begin{equation}
a_{c} = \frac{e^{\mathbf{A}_{c}\mathbf{z}}}{e^{\mathbf{A}_{c}\mathbf{z}} + e^{\mathbf{B}_{c}\mathbf{z}}}, ~b_{c} = \frac{e^{\mathbf{B}_{c}\mathbf{z}}}{e^{\mathbf{A}_{c}\mathbf{z}} + e^{\mathbf{B}_{c}\mathbf{z}}}.
\end{equation}
Within the framework where \(\mathbf{A}\) and \(\mathbf{B}\) are elements of \(\mathbb{R}^{C \times d}\), \(\mathbf{A}_{c} \in \mathbb{R}^{1 \times d}\) signifies the \(\mathit{c}\)-th element of \(\mathbf{A}\), while \(a_{c}\) denotes the \(\mathit{c}\)-th element of \(\mathbf{a}\); a similar notation applies to \(\mathbf{B}_{c}\) and \(b_{c}\). In a dual-branch configuration, the presence of matrix \(\mathbf{B}\) becomes superfluous as a linear relationship \({a_{c} + b}_{c} = 1\) holds true. Consequently, the feature map \(\mathbf{V}\) is synthesized by aggregating weighted kernels, where \(\mathbf{V} = \left\lbrack {{\mathbf{V}_{1},\mathbf{V}}_{2},\ldots,\mathbf{V}}_{c} \right\rbrack\), \(\mathbf{V}_{c} \in \mathbb{R}^{\mathit{H^{'}} \times \mathit{W^{'}}}\):
\begin{equation}
\mathbf{V}_{c} = a_{c} \cdot {\tilde{\mathbf{U}}}_{c} + b_{c} \cdot {\hat{\mathbf{U}}}_{c}.
\end{equation}
We construct an attention map that captures important spatial information using a Sigmoid activation function \(\sigma\) and a \(1 \times 1\) convolution \(\mathcal{F}_{1 \times 1}\). The enhanced attention feature map \(\mathbf{U}^{\prime}\) is subsequently generated by element-wise multiplication of this attention map with the original input feature map \(\mathbf{U}\):
\begin{equation}
\mathbf{U}^{\prime} = \mathbf{U} \cdot \sigma\left( {\mathcal{F}_{1 \times 1}\left( \mathbf{U} \right)} \right).
\end{equation}

We obtain the output feature map \(\mathbf{X^{'}}\) of the SCAN module by combining \(\mathbf{X}\) with the enhanced attention feature map \(\mathbf{U}^{\prime}\) and selected feature map \(\mathbf{V}\):
\begin{equation}
\mathbf{X^{'}} = \mathbf{X} + \mathbf{U}^{\prime} + \mathbf{V}.
\end{equation}

The transformer encoder \cite{2017Attention} consists of alternating layers of multi-headed self-attention (MSA) and Multilayer Perceptron (MLP) blocks. Layernorm (LN) is applied before every block, and residual connections after every block. The MLP contains two layers with a GELU non-linearity. The transformer is processed as follows:
\begin{equation}
\left\{ \begin{array}{l}
{\mathbf{X^{''}} = MSA\left( {LN(\mathbf{X^{'}})} \right) + \mathbf{X^{'}}} \\
{\mathbf{X}_{out} = MLP\left( {LN(\mathbf{X^{''}})} \right) + \mathbf{X^{''}}}
\end{array} \right.
\end{equation}

The TransScan module is designed to enable neurons to selectively focus and extract features from different receptive fields. It can better understand complex images and improve the model's ability to process vision tasks.

\subsubsection{Distillation Loss} 
Common distillation loss functions encompass \(\mathit{l}_{1}\)-norm, \(\mathit{l}_{2}\)-norm, cross-entropy, MSE and KL divergence \cite{gou2021knowledge}. In this study, we utilize a blend of MSE and KL divergence for the loss function. KL divergence is commonly employed to quantify the similarity between two probability distributions. It assists the student model in acquiring distributional insights from the teacher model, thereby enhancing the retention of the teacher model's knowledge. The MSE loss aids the student model in directly assimilating the log probability distribution information from the teacher model, bypassing the requirement for indirect acquisition via probability distribution softening. By integrating the use of MSE and KL divergence, we can fully utilize their respective strengths to improve the effectiveness of knowledge distillation.
\begin{equation}
\mathcal{L}_{FD} = MSE\left( {F_{t}, F_{s}} \right) + KL\left( {F_{t},F_{s}} \right).
\end{equation}

\subsection{Fine-tuning on Distilled Model}
In the fine-tuning process, we classify the fine-tuning into slide-level and patch-level based on the differences between histopathology images and other images. As shown in Fig. \ref{fig:framework}, the whole process of slide-level image data processing and distillation fine-tuning is given. The patch-level fine-tuning is relatively straightforward. The process involves adding a basic fully connected classification head to facilitate end-to-end fine-tuning and ultimately achieve the desired classification results.

For slide-level pathology images, it is typically necessary to use MIL to synthesize a series of instances into a bag sample for classification. Due to computational limitations,  the slide-level gradients cannot be backpropagated in parallel to a feature encoder with more than 10k instances of a bag.  Therefore, during the fine-tuning process of slide-level pathology images, it is necessary to sample the instances for each WSI, corresponding to Stage 3 in the EFCM framework as depicted in Fig. \ref{fig:framework}. After the instance sampling is completed, a small number of instance samples are used to perform end-to-end fine-tuning of the distilled student model \cite{li2023task}.

We employs a progressive approach to compare three different strategies for evaluating the performance of distillation models. These methods comprise Reuse CLAM, Retrain CLAM, and End2end Train CLAM. CLAM is a weakly supervised learning technique that utilizes an attention mechanism. It collectively identifies a sequence of instances as bag samples to achieve accurate slide classification using MIL.

In the Reuse CLAM strategy, the student model acquired through distillation employs the CLAM classification head of the teacher model. In the Retrain CLAM strategy, the distilled student model needs to be frozen,  and the CLAM classification head retrained during the fine-tuning process. The ETC strategy corresponds to the Stage 4 in the EFCM framework. The distilled student model and the CLAM classification head undergo end-to-end training using the selected patches to refine the distilled model. Then, the fine-tuned model parameters are frozen to act as a feature extractor, while a new CLAM classification head is trained to evaluate the performance of the fine-tuned model.

\section{Experiments}
\label{sec:exp}
The purpose of this experiment is to validate the significant performance improvement and efficiency gains of the EFCM framework for slide-level classification of pathology images. We compare the FPD method with the traditional feature distillation method to validate the effectiveness of the TransScan module in distilled fine-tuning. In addition, we apply the EFCM framework to generalize verification in patch-level tasks. Finally, the generalization of the TransScan module to pre-training and parameter-efficient fine-tuning also proves to bring some improvement.

\subsection{Experimental Details}
Our experimental subjects comprise three large models in the medical domain: RETFound for retina \cite{zhou2023foundation}, MRM for chest X-ray \cite{zhou2023advancing}, and BROW for histopathology \cite{wu2023brow}. These models address crucial tasks across various medical domains. 

\subsubsection{Dataset Details}
Our experiment consists of a total of 11 datasets, all of which are downstream task datasets for large models and are not present in the training data of the large model. As shown in Table \ref{tab:datasets}, this table summarizes the dataset information in different medical fields such as retina, chest X-ray and histopathology. It includes details on classes, disease categories, and the distribution of training, validation, and test data. All downstream datasets are publicly accessible and available online.

To enhance the diversity of the retinal images, a set of augmentation procedures is executed, with detailed parameter configurations delineated in Table \ref{tab:augment}. These augmentation processes contribute to a broader and enriched dataset of images.

\begin{table}[tbp]
\caption{Overview of downstream task datasets for three large models.} \label{tab:datasets}
\begin{adjustbox}{center}
\begin{threeparttable}
\scalebox{0.87}{
\begin{tabular}{lcrr}
\toprule
\multicolumn{1}{l}{Datasets} & Classes & Disease Category     & \begin{tabular}[c]{@{}c@{}}Data split \\      train$\mathbf{/}$val$\mathbf{/}$test\end{tabular} \\ \midrule
IDRiD  \cite{porwal2020idrid}  & 5  & \multirow{3}{*}{Diabetic retinopathy} & 329$\mathbf{/}$84$\mathbf{/}$ 103 \\
MESSIDOR-2 \cite{decenciere2014feedback} & 5  &                      & 972$\mathbf{/}$246$\mathbf{/}$526       \\
APTOS  \cite{karthik2019aptos} & 5  &                      & 2,048$\mathbf{/}$514$\mathbf{/}$1,100     \\
\cmidrule{3-4}
PAPILA  \cite{kovalyk2022papila}  & 3  & \multirow{2}{*}{Glaucoma}             & 312$\mathbf{/}$79$\mathbf{/}$98 \\
Glaucoma Fundus \cite{ahn2018deep} & 3  &                      & 861$\mathbf{/}$218$\mathbf{/}$465       \\
\cmidrule{1-4}
NIH ChestX-ray \cite{wang2017chestx} & 14 & \multirow{3}{*}{Pneumonia} & 78,468$\mathbf{/}$11,219$\mathbf{/}$22,433 \\
CheXpert  \cite{irvin2019chexpert}  & 5  &                      & 218,414$\mathbf{/}$5,000$\mathbf{/}$234   \\
RSNA Pneumonia \cite{shih2019augmenting} & 2  &                      &25,184$\mathbf{/}$1,500$\mathbf{/}$3,000    \\
\cmidrule{1-4}
TCGA-NSCLC  \cite{grossman2016toward} & 2  & Lung Cancer          & 800$\mathbf{/}$200$\mathbf{/}$200       \\
PANDA   \cite{bulten2022artificial} & 2  & Prostate Cancer      & 7,431$\mathbf{/}$1,061$\mathbf{/}$2,123    \\
TCGA-BRCA  \cite{cancer2012comprehensive}  & 2  & Breast Cancer        & 779$\mathbf{/}$97$\mathbf{/}$97         \\
\midrule
\bottomrule                                               
\end{tabular}
}
\end{threeparttable}
\end{adjustbox}
\end{table}

\begin{table}[tbp]
\centering 
\caption{The overview of the image augmentation methods and corresponding parameters.}
\label{tab:augment}
\begin{adjustbox}{center}
\begin{threeparttable}
\scalebox{1.1}{
    \begin{tabular}{lc}
        \toprule
        \multicolumn{1}{c}{Augmentation method} & \multicolumn{1}{c}{Parameters} \\ 
        \cmidrule{1-2}
		Brightness       & \{0.5, 0.7, 1.3, 1.5\}             \\
		Contrast         & \{0.5, 0.8, 1.2, 1.5\}             \\
		Color            & \{0.5, 0.8, 1.2, 1.5\}             \\
		Sharpness        & \{0.5, 0.8, 1.2, 1.5\}             \\
		Gaussian Blur    & \{1, 2, 3\}                        \\
		Flip             & \{L\_R, T\_B\}                     \\
		Rotate           & \{-45°, -30°, -15°, 15°, 30°, 45°\} \\
		Noise            & \{0.05, 0.1\}                       \\
        \midrule
        \bottomrule
    \end{tabular}}
\end{threeparttable}
\end{adjustbox}
\end{table}

\subsubsection{Distillation Training Details}
In distillation training, we follow the standard practices for data augmentation by resizing input images to 224 \(\times\) 224 and normalizing them through practical mean channel subtraction. We choose the AdamW optimizer \cite{kingma2014adam} to adjust model parameters using the following settings: a learning rate of 1e-4, beta values of (0.9, 0.999), weight decay of 1e-2, and epsilon of 1e-8. For learning rate adjustment, we utilize the CosineAnnealingLR \cite{loshchilov2016sgdr} as a learning rate scheduler with a warm-up step of 200, causing the learning rate to increase linearly from 0 to the initial setting of 1e-4 during the warm-up phase, followed by cosine annealing to adjust the learning rate smoothly throughout training. In addition, we use a batch size of 64 for parallel data processing to optimize computational efficiency. The AdamW optimizer updates model parameters based on the specified learning rate, weight decay, and other parameter values.

\subsubsection{Fine-tuning Implementation}
For fine-tuning on distilled model, we initially initialize the model with the weights trained through distillation. In the fine-tuning process of the student model of FPD, the shallow ResNet50 parameters remain frozen, while other parameters are fine-tuned at a lower learning rate, typically set to 1e-5. In the fine-tuning of VFD, the learning rate is also set to 1e-5. The learning rate of the classifier head is usually set to 5e-3 when the classification task is performed.

During the fine-tuning stage, the images are randomly cropped to 224 \(\times\) 224, as well as random horizontal flipping and standardization. The training process employs a batch size of 16, and we adopt the AdamW optimizer to adjust the model's parameters, set an appropriate learning rate, and apply weight decay. To mitigate overfitting, we incorporate label smoothing to soften the true labels of the training data and adjust the output distribution. Following each epoch, the model is evaluated on the validation set, and the weights of the model with the highest AUC on the validation set are saved as checkpoints for both internal and external evaluations.

In the case of full-parameter fine-tuning, no parameters need to be frozen. However, during parameter-efficient fine-tuning, specific parameters need to be adjusted. In contrast to distillation fine-tuning model, both full-parameter fine-tuning and parameter-efficient fine-tuning, when combined with the classification head, utilize the same learning rate, typically set at 5e-3. Other settings can be referenced from the parameters used in the distillation fine-tuning process.

\begin{figure}[tbp]
    \centering
    \includegraphics[scale=.6]{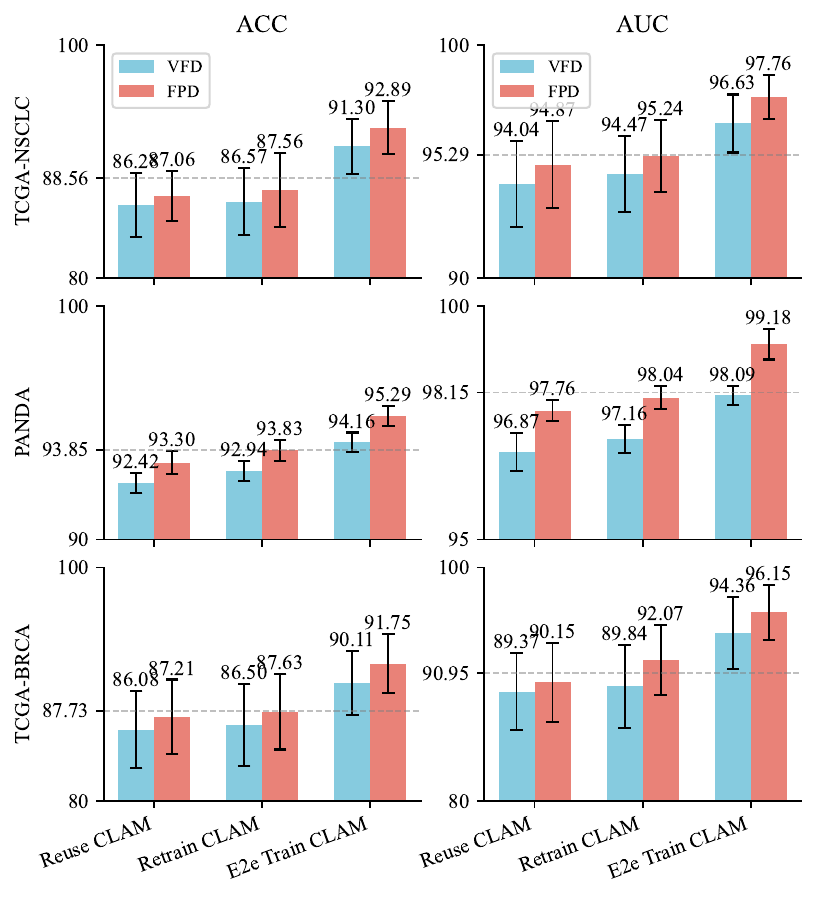}
    \caption{The performance of two distillation models is compared on pathology image datasets using three fine-tuning strategies. The VFD method is represented by sky blue, the FPD method by salmon, and the metrics of the large model on each dataset are depicted by a light gray dashed line.}
    \label{fig:e2e}
\end{figure}

\subsection{Study of Distillation Fine-tuning}
The distillation fine-tuning process consists of two stages. This section focuses on the distillation fine-tuning method for slide-level pathology images, the process of which is shown in Fig. \ref{fig:framework}. We further generalize the distillation fine-tuning method to patch-level downstream tasks.

\subsubsection{Slide-level Distillation Fine-tuning}
The results of distillation fine-tuning across three strategies applied to the pathology image dataset are illustrated in Fig. \ref{fig:e2e}. A marginal increase of approximately 0.5\% is evident when comparing the Retrain CLAM with the Reuse CLAM. Notably, the ETC strategy exhibits superior performance in both ACC and AUC. The  ETC strategy can enhance ACC by 5.83\%, 1.99\%, and 4.54\%  compared to the Reuse CLAM on three distinct datasets. Particularly, the TCGA-BRCA dataset showcases a remarkable 6\% improvement in AUC. In summary, the ETC strategy is a promising approach for improving the accuracy and efficiency of models, particularly in tasks involving slide-level image recognition.

\begin{figure}[tbp]
    \centering
    \includegraphics[width=\linewidth]{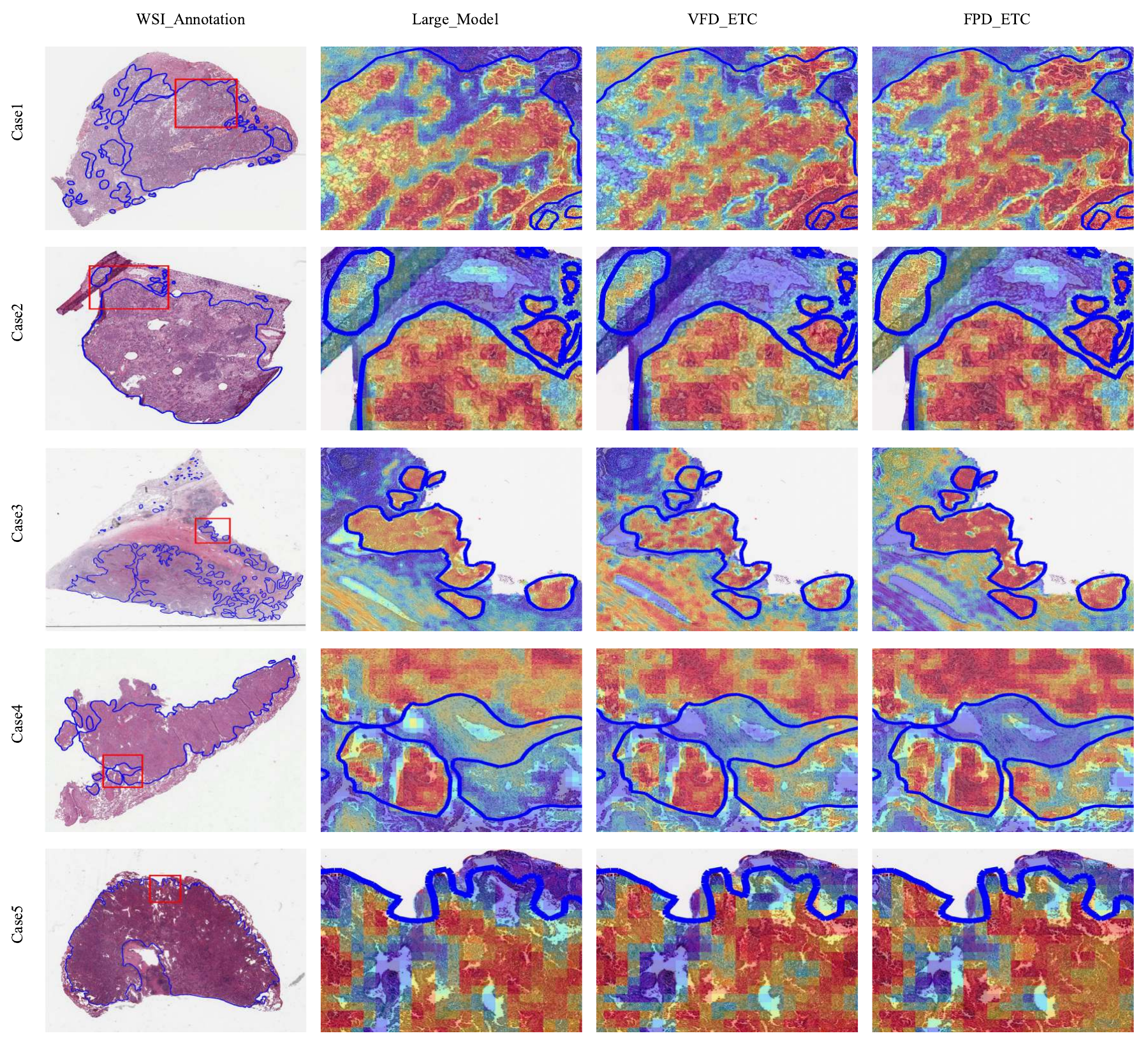}
    \caption{Visualization of the results for some cases. These cases are from the TCGA-NSCLC dataset. The first column of images represents the real situation of the lesion area marked with a blue line, with red rectangles indicating local ROIs highlighting the boundary between the tumor and normal tissue. Columns 2 to 4 display the ROIs of the large model, VFD, and FPD methods predicting the attentional heatmap. Warmer colors in the attentional heatmap indicate a higher probability of estimating tumor tissue.}
    \label{fig:heatmap}
\end{figure}

In addition, combining these three strategies we compare VFD and FPD. The results reveal that the distilled student model of FPD method outperforms the VFD method across all fine-tuning strategies. Particularly, we observe a more substantial performance boost of up to 5.2\% on the TCGA-BRCA dataset, surpassing the performance delivered by the BROW model. On the TCGA-BRCA dataset, Li \textit{et al.} \cite{li2023task} achieve an AUC of 93.5\% by fine-tuning the ResNet50 model. We fine-tune on distilled ResNet50 model and attain an AUC of 94.36\%, resulting in a performance improvement of 0.86\%. Furthermore, our FPD\_ETC method demonstrates a significant increase in AUC of 2.65\% compared to the method in \cite{li2023task}. These results validate the effectiveness of the distillation techniques and highlight the advantages of feature projection distillation fine-tuning.

We visualize and analyze some cases, as shown in Fig. \ref{fig:heatmap}. The visualization results compare the attentional heatmaps corresponding to different model classifications, including the proposed FPD\_ETC method and VFD\_ETC, as well as the large model approach. We all employ the identical feature aggregation scheme as the CLAM method. These attentional heatmaps are generated based on the importance of each sub-region in the classification process. Our FPD\_ETC method generates highly accurate heatmaps of localized tumors that closely correspond to the ground truth.

Performance comparison of the proposed FPD\_ETC with state-of-the-art methods on the TCGA-NSCLC dataset. As shown in Table \ref{tab:sota}, our FPD\_ETC method outperforms current state-of-the-art methods. Specifically, compared with the top-performing methods, MSPT and LKA, our method achieves a 1.34\% increase in AUC and a 0.99\% improvement in ACC for binary classification on the TCGA-NSCLC dataset. On the PANDA and TCGA-BRCA datasets, our proposed FPD\_ETC method is compared with state-of-the-art methods, and the results show that it also performs the best in terms of AUC, as detailed in Tables \ref{tab:panda_sota} and \ref{tab:brac_sota}.

\begin{table}[tbp]
\centering
\caption{Performance comparison with the state-of-the-art methods on the TCGA-NSCLC dataset.}
\label{tab:sota}
\begin{adjustbox}{center}
\begin{threeparttable}
\scalebox{0.66}{
    \begin{tabular}{ccccccccccc}
        \toprule
        \multirow{2}{*}{Method} & \multirow{1}{*}{SCL-WC} & \multirow{1}{*}{SRCL} & \multirow{1}{*}{GTP} & \multirow{1}{*}{IGT} & \multirow{1}{*}{CaMIL} & \multirow{1}{*}{ReMix} & \multirow{1}{*}{MSPT} & \multirow{1}{*}{BCL} & \multirow{1}{*}{LKA} & \multirow{2}{*}{FPD\_ETC} \\
              &  \cite{wang2022scl} &  \cite{wang2022transformer} &  \cite{zheng2022graph}  &  \cite{shi2024integrative}  &  \cite{chen2024camil}  &   \cite{yang2022remix}  &  \cite{ding2023multi}  &   \cite{yu2023bayesian}   &  \cite{yao2023self}  &     \\
        \cmidrule{1-11}
        ACC   & -      & 91.2 & 90.5 & 91.6 & 90.0  & 91.67 & 92.89 & 90.8  & 91.9  & \textbf{92.89} \\
        AUC   & 97.1   & 97.3 & 95.8 & 96.7 & 95.64 & 95.09 & 96.22 & 96.0  & 97.54 & \textbf{97.56} \\
        \midrule
        \bottomrule
    \end{tabular}}
\end{threeparttable}
\end{adjustbox}
\end{table}

\begin{table}[tbp]
\centering
\caption{Performance comparison with the state-of-the-art methods on the PANDA dataset.} 
\label{tab:panda_sota}
\begin{adjustbox}{center}
\begin{threeparttable}
\scalebox{0.95}{
    \begin{tabular}{cccccc}
        \toprule
        \multirow{2}{*}{Method} & \multirow{1}{*}{AB-MIL} & \multirow{1}{*}{SCL-WC} & \multirow{1}{*}{FederatedHN} & \multirow{1}{*}{IS-MIL} & \multirow{2}{*}{FPD\_ETC} \\
              &  \cite{ilse2018attention} &  \cite{wang2022scl} &  \cite{lin2023federated}  &  \cite{yang2023devil}  &     \\
        \cmidrule{1-6}
        AUC   & 95.14  & 97.53  & 95.7  & 98.7  & \textbf{99.18} \\
        \midrule
        \bottomrule
    \end{tabular}}
\end{threeparttable}
\end{adjustbox}
\end{table} 
  
\begin{table}[!t] 
\centering 
\caption{Performance comparison with the state-of-the-art methods on the TCGA-BRCA dataset.}
\label{tab:brac_sota}
\begin{adjustbox}{center}
\begin{threeparttable}
\scalebox{0.9}{
    \begin{tabular}{cccccc}
        \toprule
        \multirow{2}{*}{Method} & \multirow{1}{*}{SAMPLER} & \multirow{1}{*}{FT+Mean-pool} & \multirow{1}{*}{Long-MIL} & \multirow{1}{*}{BEPH} & \multirow{2}{*}{FPD\_ETC} \\
              &  \cite{mukashyaka2024sampler} &  \cite{li2023task} &  \cite{zheng2022graph}  &  \cite{yu2024foundation}  &     \\
        \cmidrule{1-6}
        AUC   & 91.1   & 95.2 & 94.6 & 94.6  & \textbf{96.15} \\
        \midrule
        \bottomrule
    \end{tabular}}
\end{threeparttable}
\end{adjustbox}
\end{table}

\begin{table*}[tbp] 
\centering 
\caption{Comparison of retinal image fine-tuning results: full-parameter fine-tuning and two distillation fine-tuning methods.}
\label{tab:fdm_ret}
\begin{adjustbox}{center}
\begin{threeparttable}
\scalebox{1.15}{
    \begin{tabular}{clcccccccccc}
        \toprule
        \multirow{2}{*}{Params} & \multirow{2}{*}{Method} & \multicolumn{2}{c}{IDRiD} & \multicolumn{2}{c}{APTOS} & \multicolumn{2}{c}{MESSIDOR-2} & \multicolumn{2}{c}{PAPILA} & \multicolumn{2}{c}{Glaucoma Fundus} \\
        \cmidrule(lr){3-4} \cmidrule(lr){5-6} \cmidrule(lr){7-8} \cmidrule(lr){9-10} \cmidrule(lr){11-12}
        (M)   &              & ACC  & AUC & ACC & AUC & ACC & AUC  & ACC & AUC & ACC & AUC  \\
        \cmidrule{1-12}
        303.31 & All Finetune & 0.8155 & 0.8266 & 0.9259 & 0.9473 & 0.9097 & 0.8783 & 0.8827 & 0.8551 & 0.9086 & 0.9495  \\
        \cmidrule{1-12}
        9.6    & VFD           & 0.818  & 0.7781 & 0.9112  & 0.9318 & 0.8924 & 0.8541 & 0.8571 & 0.7737 & 0.8624 & 0.9212  \\
        7.88   & FPD         & 0.8665 & 0.8401 & 0.9168 & 0.9374  & 0.9069 & 0.8672   & 0.8776 & 0.8353 & 0.8934 & 0.9354  \\
        \midrule
        \bottomrule
    \end{tabular}}
\end{threeparttable}
\end{adjustbox}
\end{table*}

In addition, we also perform model distillation training on different datasets and transfer it to other datasets to perform fine-tuning operations to evaluate the generalization ability of the distilled model.

\begin{figure}[tbp]
    \centering
    \includegraphics[width=\linewidth]{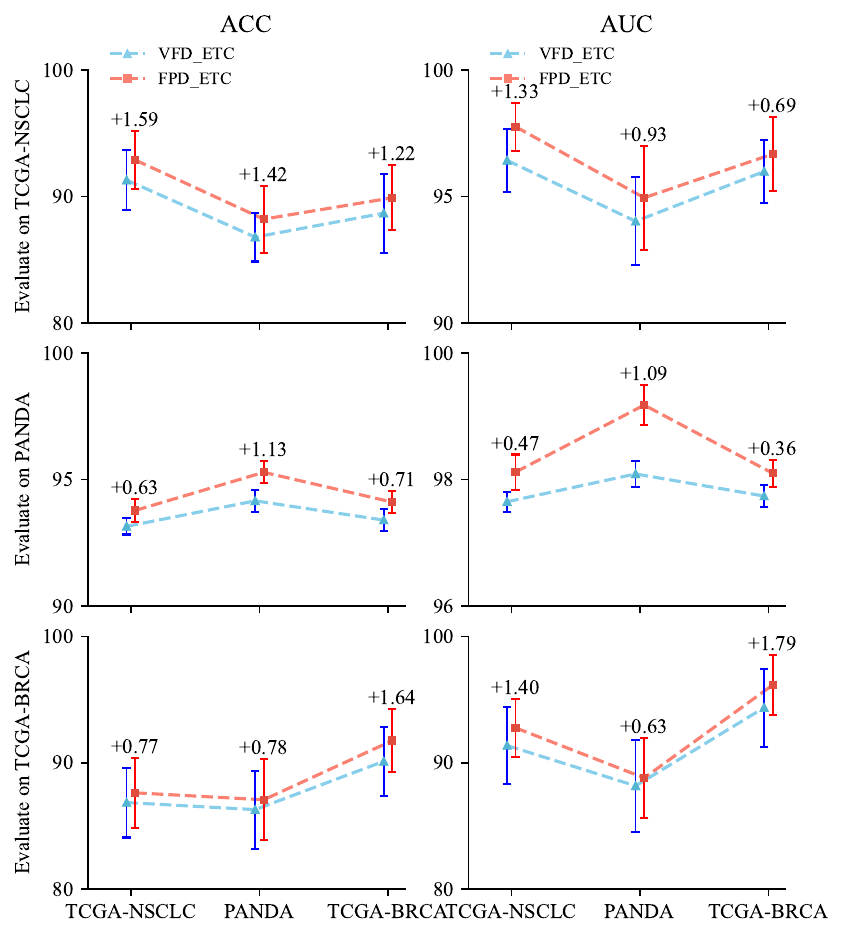}
    \caption{Evaluate the transferability of distillation models using the ETC fine-tuning strategy. The assessment of distillation model transferability across datasets is evaluated through fine-tuning. The VFD\_ETC method is represented by sky blue, and the FPD\_ETC method is represented by salmon.}
    \label{fig:data_transfer}
\end{figure}

According to the results shown in Fig. \ref{fig:data_transfer}, the fine-tuning performance is best when the distillation-trained dataset and the fine-tuned dataset are identical. This performance is superior to fine-tuning using the transferred distillation model. The fine-tuning performance of the FPD\_ETC method is better than that of the VFD\_ETC method in both ACC and AUC, with an improvement ranging from 0.36\% to 1.42\%. In addition, the findings suggest that models obtained by distillation on the PANDA dataset tend to perform poorer when transferred to the other two datasets for fine-tuning. Similarly, models obtained by distillation on the other two datasets also exhibit mediocre performance when fine-tuned on the PANDA dataset. This difference in performance may be due to the different feature distributions between the different datasets.

This finding implies that researchers need to consider the feature compatibility among different datasets, along with the feasibility of transfer learning when performing transfer fine-tuning of the distilled models. In practical applications, by rationally exploiting the similarities among the features of the datasets, we can effectively guide the transfer learning of the model.

\subsubsection{Patch-level Distillation Fine-tuning}
Retina and chest X-ray are not as large as pathology images and require only the addition of a classification head for end-to-end patch-level fine-tuning. The patch-level fine-tuning experiments are conducted on retina and chest X-ray datasets using three distinct models: a VFD model, an FPD model, and a large model.

\begin{table}[tbp]
\centering 
\caption{Comparison of chest X-ray image fine-tuning results: full-parameter fine-tuning and two distillation fine-tuning methods.}
\label{tab:fdm_mrm}
\begin{adjustbox}{center}
\begin{threeparttable}
\scalebox{0.92}{
    \begin{tabular}{lcccccc}
        \toprule
        \multirow{2}{*}{Method} & \multicolumn{2}{c}{NIH ChestX-ray} & \multicolumn{2}{c}{CheXpert} & \multicolumn{2}{c}{RSNA Pneumonia} \\
        \cmidrule(lr){2-3} \cmidrule(lr){4-5} \cmidrule(lr){6-7}
        \multicolumn{1}{c}{}   & ACC  & AUC   & ACC   & AUC  & ACC   & AUC   \\
        \cmidrule{1-7}
        All Finetune    & 0.949    & 0.859    & 0.8197   & 0.887    & 0.8183    & 0.9324    \\
        \cmidrule{1-7}
        VFD             & 0.9384   & 0.824    & 0.8011   & 0.8543   & 0.7963    & 0.9114     \\
        FPD             & 0.9483   & 0.8367   & 0.8086   & 0.8752   & 0.8407    & 0.9353     \\ 
        \midrule
        \bottomrule
    \end{tabular}}
\end{threeparttable}
\end{adjustbox}
\end{table}

The fine-tuning experiments for retinal and chest X-ray images follow a similar design, with the distilled models being fine-tuned on respective datasets. We assess the performance of each model by evaluating metrics such as  ACC and AUC, aiming to determine their effectiveness in classifying retinal diseases and detecting abnormalities in chest X-ray images. The analysis of the parameter count of each model indicates that the distilled model by the FPD method is the most practical for real-world deployment due to its lower parameter count.

Tables \ref{tab:fdm_ret} and \ref{tab:fdm_mrm} present the results of fine-tuning on retinal and chest X-ray image datasets. The FPD method enhances the performance of the large model by 5.1\% and 2.24\% in ACC on the IDRiD and RSNA Pneumonia datasets, respectively. On other datasets, this model demonstrates comparable performance to full-parameter fine-tuning of the large model. Compared to the VFD model, the FPD model exhibits superior performance in various downstream classification tasks. It is noteworthy that the FPD model comprises only 7.88 million parameters, which is nearly one-fortieth of the parameter count in the large model and 1.72 million fewer parameters than the VFD model. These results demonstrate the accuracy and efficiency of the FPD fine-tuning method.

\subsection{Ablation Study}
In this section, our primary goal is to evaluate the impact of different loss functions and model architectures on the performance of distillation fine-tuning. We also aim to explore what hyperparameters can constitute a good TransScan module. To ensure the generalizability and reliability of our findings, we conduct all experiments on the IDRiD dataset. Our research employs a systematic approach to experimentation, replacing components or parameters of the model step-by-step to observe the effect on overall performance. 

We experimentally explore the effect of different distillation losses on distillation fine-tuning performance, and the results are shown in Table \ref{tab:loss}. Our ablation study reveals that employing a combination of MSE and KL divergence loss terms can yield superior performance. Specifically, the MSE loss aims to minimize the absolute error between predicted and target values, while the KL loss aims to reduce the distributional disparity between predicted and target values. By integrating these two losses, we can optimize these critical aspects concurrently, thereby enhancing the performance of distillation fine-tuning.

We compare the effect of different model architectures on distillation, as shown in Table \ref{tab:loss}. Our results show that the hybrid student model combining CNN and Transformer improves knowledge distillation, but only improves ACC by 0.97\%. Noteworthy is the observation that the FPD method with TransScan module achieves a significant increase in performance, marking up to 4.85\% and 6.2\% improvement in ACC and AUC, respectively. The introduction of the TransScan module provides a remarkable improvement in model performance.

\begin{table}[tbp]
\centering 
\caption{Effect of different distillation losses and model architectures on distillation fine-tuning performance.}
\label{tab:loss}
\begin{adjustbox}{center}
\begin{threeparttable}
\scalebox{1.0}{
    \begin{tabular}{lcccccc}
        \toprule
        \multirow{2}{*}{Architecture} & \multicolumn{2}{c}{MSE} & \multicolumn{2}{c}{KL} & \multicolumn{2}{c}{MSE + KL} \\
        \cmidrule(lr){2-3} \cmidrule(lr){4-5} \cmidrule(lr){6-7}
        \multicolumn{1}{c}{}   & ACC  & AUC   & ACC   & AUC  & ACC   & AUC   \\
        \cmidrule{1-7}
        CNN         &  78.88  &  73.70  &  79.61  &  74.39  &  81.8   & 77.81  \\
        Transformer &  79.21  &  79.04  &  79.37  &  76.51  &  79.85  & 76.9   \\
        FPD\_noSCAN &  81.55  &  77.42  &  82.77  &  74.15  &  82.77  & 78.32   \\
        FPD         &  84.17  &  83.72  & 85.78   &  83.63  &  \textbf{86.65}  & \textbf{84.01}   \\
        \midrule
        \bottomrule
    \end{tabular}}
\end{threeparttable}
\end{adjustbox}
\end{table}

\begin{table}[tbp]
\centering 
\caption{Comparison of hyperparameter settings and performance that affect the SCAN structure.}
\label{tab:gd}
\begin{adjustbox}{center}
\begin{threeparttable}
\scalebox{1.15}{
    \begin{tabular}{ccccccc}
        \toprule
        \multirow{2}{*}{G} & \multicolumn{2}{c}{\(\mathit{d}\)=16} & \multicolumn{2}{c}{\(\mathit{d}\)=32} & \multicolumn{2}{c}{\(\mathit{d}\)=64} \\
        \cmidrule(lr){2-3} \cmidrule(lr){4-5} \cmidrule(lr){6-7}
        \multicolumn{1}{c}{}   & ACC  & AUC   & ACC   & AUC  & ACC   & AUC   \\
        \cmidrule{1-7}
        16  & 83.50	& 82.74	& 81.07	& 81.18	 & 79.85	& 81.73 \\
        32  & 83.25	& 83.83	& \textbf{86.65} & \underline{84.01}	& 83.98	& 82.71 \\
        64  & 83.25	& 81.98	& \underline{84.95}	 & \textbf{84.43}	& 81.31	& 83.59 \\
        \midrule
        \bottomrule
    \end{tabular}}
\end{threeparttable}
\end{adjustbox}
\end{table}

We further investigate the settings of hyperparameters G and \(\mathit{d}\) in the SCAN structure. The meanings represented by the hyperparameters G and \(\mathit{d}\) can be found in Section \ref{sec:scan}. The comparison results in Table \ref{tab:gd} show that appropriate hyperparameter settings can improve model performance. Overall, the optimal model performance is obtained when both hyperparameters G and \(\mathit{d}\) are set to 32.

\begin{table}[tbp]
\centering 
\caption{Effect of the parameters ``Depth'' and ``dim'' of the TransScan module in the FPD method on the distillation fine-tuning performance and the number of parameters.}
\label{tab:dim}
\begin{adjustbox}{center}
\begin{threeparttable}
\scalebox{0.76}{
    \begin{tabular}{cccccccccc}
        \toprule
        \multirow{2}{*}{Depths} & \multicolumn{3}{c}{dim=192} & \multicolumn{3}{c}{dim=384} & \multicolumn{3}{c}{dim=576} \\
        \cmidrule(lr){2-4} \cmidrule(lr){5-7} \cmidrule(lr){8-10}
        \multicolumn{1}{c}{}   & Params &  ACC  & AUC   &  Params & ACC   & AUC  &  Params & ACC   & AUC   \\
        \cmidrule{1-10}
            2 & 2.19M & 83.50 & 83.83 & 5.99M & 83.25 & 83.16 & 11.65M & 83.98 & 83.60 \\
            3 & 2.67M & 82.33 & 83.29 & 7.88M & \textbf{86.65} & 84.01 & 15.89M & 83.50 & 81.92 \\
            4 & 3.16M & \underline{84.95} & \underline{84.04} & 9.79M & 83.01 & \textbf{84.34} & 20.12M & 82.52 & 79.23 \\  
        \midrule
        \bottomrule
    \end{tabular}
}
\end{threeparttable}
\end{adjustbox}
\end{table}

We also explore how the feature transformation dimension (referred to as ``dim") and the depth (referred to as ``Depth") of the concatenated TransScan module affect the performance of the models, as shown in Table \ref{tab:dim}. As both ``Depth" and ``dim" increase, the number of model parameters increases accordingly, requiring careful consideration when balancing model performance and computational efficiency. We find that setting ``Depth" to 3 and ``dim" to 384 achieves the optimal balance between performance and computational efficiency.

\begin{table*}[tbp]
\centering 
\caption{Performance comparison of full-parameter fine-tuning and parameter-efficient fine-tuning methods on retina datasets.}
\label{tab:transscan}
\begin{adjustbox}{center}
\begin{threeparttable}
\scalebox{1.15}{
    \begin{tabular}{clcccccccccc}
        \toprule
        \multirow{2}{*}{Params} & \multirow{2}{*}{Method} & \multicolumn{2}{c}{IDRiD}         & \multicolumn{2}{c}{APTOS}         & \multicolumn{2}{c}{MESSIDOR-2}   & \multicolumn{2}{c}{PAPILA}        & \multicolumn{2}{c}{Glaucoma Fundus} \\ 
        \cmidrule(lr){3-4} \cmidrule(lr){5-6} \cmidrule(lr){7-8} \cmidrule(lr){9-10} \cmidrule(lr){11-12}
        (M) & & ACC & AUC & ACC & AUC & ACC & AUC & ACC & AUC & ACC & AUC \\
        \cmidrule{1-12}
        303.31 & All Finetune & 0.8155 & 0.8266 & 0.9259 & 0.9473 & 0.9097 & 0.8783 & 0.8827 & 0.8551 & 0.9086 & 0.9495       \\
        \cmidrule{1-12}
        +3.18 & Adapter      & 0.8083  & \textbf{0.8236} & \underline{0.9239}   & \textbf{0.9475} & \underline{0.9064}  & 0.8761 & 0.8724 & \underline{0.8611} & \textbf{0.9272}  & 0.9574 \\
        +3.15 & LoRA   & \textbf{0.8325} & 0.8056  & 0.9214 & \underline{0.9473} & 0.894 & \underline{0.8801} & \underline{0.8776} & 0.8414 & 0.9238  & \underline{0.9577}  \\
        +2.12 & AdaptScan      & \underline{0.8228}    & \underline{0.8165}  & \textbf{0.9255} & 0.9456 & \textbf{0.9097} & \textbf{0.8808} & \textbf{0.8929} & \textbf{0.8716} & \underline{0.9247}  & \textbf{0.9585} \\  
        \midrule
        \bottomrule
    \end{tabular}}
\end{threeparttable}
\end{adjustbox}
\end{table*}

\subsection{Analysis of Model Efficiency}
Optimizing model efficiency is critical in the rapidly evolving field of artificial intelligence, especially under resource constraints or stringent inference speed requirements. Researchers aim to achieve an optimal balance between model performance and resource consumption by employing a variety of techniques and strategies. This section provides a thorough discussion of the benefits and limitations of several approaches, offering readers valuable insights into large model optimization.

\begin{figure}[b]
    \centering
    \includegraphics[scale=.85]{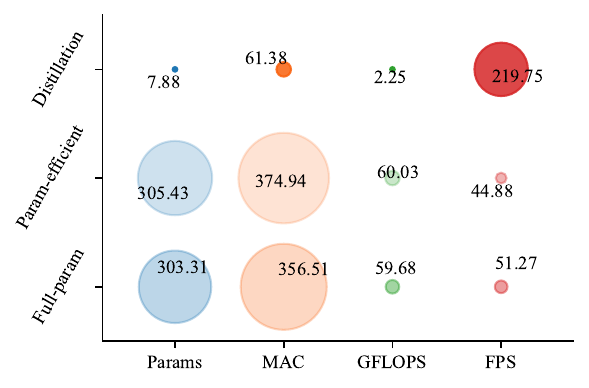}
    \caption{Comparative analysis of model efficiency for three fine-tuning methods.}
    \label{fig:efficiency}
\end{figure}

Fig. \ref{fig:efficiency} provides a comprehensive analysis of the efficiency of three methods, including full-parameter fine-tuning, parameter-efficient fine-tuning, and distillation fine-tuning. Metrics evaluated include Params, MAC, GFLOPS, and FPS, providing insight into the computational efficiency and inference speed of each method.

The parameter-efficient fine-tuning optimizes the model by updating a part of the parameters, reducing the computational load on the training stage without compromising model complexity. As shown in Fig. \ref{fig:efficiency}, parameter-efficient fine-tuning exhibits a modest rise in MAC and GFLOPS, indicating that they entail some extra computational burden during the inference phase compared to full-parameter fine-tuning. Hence, while these methods for fine-tuning have proven to be efficacious in enhancing model performance, it is essential to recognize that they will introduce additional computational intricacy that does not contribute to speedup in inference.

The distillation fine-tuning method provides significant enhancements in FPS and reductions in Params, MAC, and GFLOPS. This aligns with the objective of distillation-based fine-tuning techniques, which seek to develop lightweight models optimized for rapid inference, rendering them suitable for real-time applications requiring low latency.

\section{Applications Study of TransScan Module}
We further deeply explore the generalization application ability of the TransScan module. The TransScan module is respectively applied to model pre-training and parameter-efficient fine-tuning to explore whether the TransScan module can also bring about performance improvement.

\subsection{TransScan for Pre-training}
TransScan Module can be used in pre-training by introducing it into existing transformer architecture models and replacing the original transformer of the model. For our experiments on the CIFAR-10 dataset, we select several different models, including ViT \cite{dosovitskiy2020image}, Swin \cite{liu2021swin}, MaxViT \cite{tu2022maxvit}, Crossformer++ \cite{wang2023crossformer++}, and EVA02 \cite{fang2023eva}, and replace the transformer blocks in these models with the TransScan Module. These models are retrained on the CIFAR-10 dataset.

\begin{figure}[tbp]
    \centering
    \includegraphics[scale=.9]{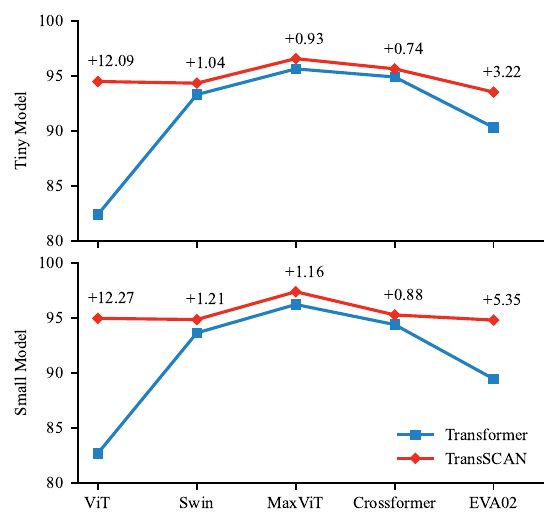}
    \caption{Evaluation of the impact of introducing SCAN structure on the performance of different models on the CIFAR-10 dataset. The plus sign indicates an improvement in model performance. 
    }
    \label{fig:pretrain}
\end{figure}

Our experimental results, shown in Fig. \ref{fig:pretrain}, demonstrate that the TransScan module can significantly improve the model performance during pre-training. By comparing the model performance before and after the replacement, we evaluate the impact of TransScan on the model in terms of improving classification accuracy. The experimental results show that TransScan achieves higher accuracy on the CIFAR-10 dataset compared to the transformer.

\subsection{TransScan for Parameter-efficient Fine-tuning}
We introduce the TransScan module in parameter-efficient fine-tuning and call this new method AdaptScan, which aims to use the SCAN structure to quickly adapt the large model to new tasks. We apply this method to five datasets of retinal images.

Table \ref{tab:transscan} presents a comparison among four methods: All Finetune, Adapter, LoRA, and AdaptScan. We employ the pre-trained large-scale retinal model RETFound, based on the ViT-Large framework, as the backbone network. In practice, the SCAN structure is not concatenated with every transformer, but rather with specific layers of transformers. For this experiment, transformers at the 12th, 14th, 16th, 18th, 20th,  and 22nd layers are selected for sequential concatenation. This strategy aims to enhance and focus on high-dimensional features,  aiding in capturing intricate nonlinear relationships and patterns within the data. Overall, the AdaptScan method demonstrates robust performance across most datasets, notably achieving higher ACC and AUC scores on the PAPILA and Glaucoma Fundus datasets. 

\section{Conclusion}
In this study, we construct a novel framework of EFCM. The framework is initially applied to slide-level pathology image classification tasks to address the limitations of traditional knowledge distillation, resulting in significant improvements in model efficiency and performance. Subsequently, we apply the method of distillation followed by fine-tuning to patch-level image tasks, successfully obtaining small models that perform comparably to large models.

In the EFCM framework, our proposed FPD method plays a crucial role, with the TransScan module being instrumental. The TransScan module enhances the model's ability to handle visual tasks by adaptively adjusting receptive fields using SCAN. Additionally, when comparing the FPD method with the VFD method, we find that the distilled models obtained through FPD preserve more knowledge from the teacher model while maximizing compression. The performance and generalization ability of these compressed models exceed those obtained through the VFD method, demonstrating the potential of our approach in distillation. 

We perform slide-level and patch-level distillation fine-tuning experiments on three large models in the medical domain. The results indicate that the FPD\_ETC method is the most effective slide-level distillation fine-tuning approach, achieving a 4.33\% increase in ACC and a 5.2\% improvement in AUC compared to the larger model in the TCGA-NSCLC and TCGA-BRCA datasets. Patch-level distillation fine-tuning enhances generalization, maintains performance, and reduces model parameters, thus enhancing its suitability for real-world deployment.

Finally, we provide a comprehensive analysis of different model fine-tuning techniques based on various metrics such as the number of parameters, MAC, GFLOPS, and FPS, which provide valuable insights for model optimization. Further research is needed to improve model efficiency and generalization and to explore the potential of TransScan in other areas.

Overall, our proposed distillation fine-tuning method shows promise in improving model efficiency and accuracy in various medical imaging tasks, and particularly excels in slide-level pathology image tasks.

\appendices

\end{document}